\documentclass[10pt,twocolumn,letterpaper]{article}

\usepackage[pagenumbers]{cvpr} % To force page numbers, e.g. for an arXiv version

\usepackage{graphicx}
\usepackage{amsmath}
\usepackage{amssymb}
\usepackage{booktabs}

\usepackage{mathtools}
\usepackage{subcaption}
\usepackage{scalefnt}
\newcommand{\usePalatino}[1]{{\fontfamily{ppl}\selectfont\scalefont{0.96} #1}}

\usepackage[pagebackref,breaklinks,colorlinks]{hyperref}

\usepackage[capitalize]{cleveref}
\crefname{section}{Sec.}{Secs.}
\Crefname{section}{Section}{Sections}
\Crefname{table}{Table}{Tables}
\crefname{table}{Tab.}{Tabs.}

\def\confName{CVPR}
\def\confYear{2023}

\begin{document}

\title{PaReprop: Fast \underline{Pa}rallelized \underline{Re}versible Back\underline{prop}agation}

\author{
	Tyler Zhu$^*$ \qquad
	Karttikeya Mangalam$^*$
 \\
	\small $^{*}$ denotes equal technical contribution  
	     \vspace{2pt}\\
	UC Berkeley
\vspace{2pt}\\
\usePalatino{tyler.zhu@berkeley.edu}
}
\maketitle

%%%%%%%%% ABSTRACT
\begin{abstract}
The growing size of datasets and deep learning models has made faster and memory-efficient training crucial. Reversible transformers have recently been introduced as an exciting new method for extremely memory-efficient training, but they come with an additional computation overhead of activation re-computation in the backpropagation phase. We present PaReprop, a fast Parallelized Reversible Backpropagation algorithm that parallelizes the additional activation re-computation overhead in reversible training with the gradient computation itself in backpropagation phase. We demonstrate the effectiveness of the proposed PaReprop algorithm through extensive benchmarking across model families (ViT, MViT, Swin and RoBERTa), data modalities (Vision \& NLP), model sizes (from small to giant), and training batch sizes. Our empirical results show that PaReprop achieves up to 20\% higher training throughput than vanilla reversible training, largely mitigating the theoretical overhead of 25\% lower throughput from activation recomputation in reversible training. 
Project page: \url{https://tylerzhu.com/pareprop}.
\end{abstract}

\begin{figure}[!tb]
    \centering
    \includegraphics[width=0.47\textwidth]{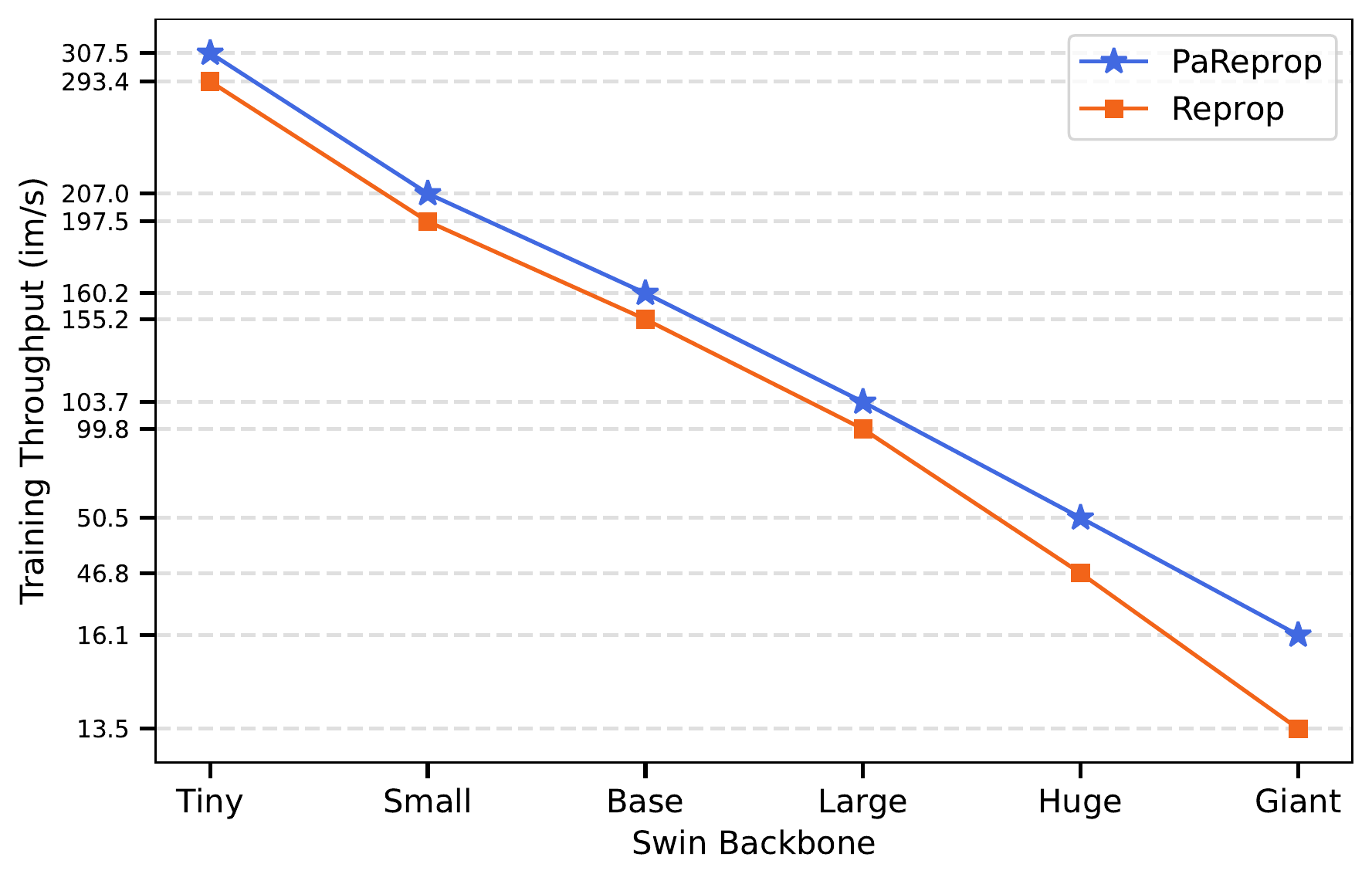}
    \caption{\textbf{PaReprop vs. Reprop for Swin Transformer}. We benchmark our proposed Parallelized Reversible backpropagation (PaReprop) algorithm against vanilla reversible backpropagation (Reprop). By exploiting parallelized kernels, PaReprop achieves up to 20\% training throughput gain without any change to the underlying computation, thereby ensuring accuracy. Notably deeper transformers enjoy better throughout improvement, a desirable quality for scaling memory-efficient reversible training.}
    \label{fig:revswin_topthr}
\end{figure}

%%%%%%%%% BODY TEXT
\section{Introduction}
\label{sec:intro}
The field of deep learning has made great strides in recent years on the back of large-scale models. 
By increasing the size of models, we achieved state of the art performances again and again on a variety of tasks, such as image recognition~\cite{vit}, language modeling~\cite{GPT3}, and speech recognition~\cite{wav2vec2}.
However, as these models become larger, it becomes equally as important to improve the efficiency of these models so that they can be deployed in real-world applications.
Models need to be able to run on smaller amounts of compute and run faster. 
When models are at the huge scale that they are today, small speedups can lead to large gains over the course of the many epochs and training \emph{weeks} that are required to train these models.

Vision transformers are able to achieve their impressive results by stacking multiple layers of transformer blocks.
However, this increase in depth is quite costly, as the number of activations that need to be stored during training increases linearly with the depth of the model.
Reversible vision transformers~\cite{revvit}, as well as other reversible architectures~\cite{gomez2017reversible,song2019mintnet,kitaev2020reformer}, is one recent result which offers a promising approach to improving the efficiency of large models by decoupling the memory needed from the depth by using reversible activations.
This allows them to maintain top performance while also requiring less memory.

The framework of reversible transformations however offers a further tradeoff where for a tiny amount of additional memory, we can improve our throughput by a signficant amount using parallelization. 
This takes advantage of the independence between the recomputations and gradient updates in the backward pass which allows them to theoretically be computed simultaneously.
From this observation, we introduce a method for parallelizing a reversible backpropagation which we call \emph{PaReprop}, or \emph{Pa}rallelized \emph{Re}versible back\emph{prop}agation.
Our method is general enough to be applied to any reversible architecture, and we demonstrate this by testing on a wide variety of reversible architectures, hardware, and memory settings, and show that we can achieve significant speedups in practice. 
In summary, we make the following contributions:
\begin{enumerate}
    \item We propose a novel method for parallelizing reversible backpropagation which is compatible with modern auto-differentiation packages like PyTorch.  
    \item Our method achieves significant speedups across a diverse set of model families, data modalities, model sizes, and training batch sizes. 
    We increase training throughput for all of our models while maintaining the same accuracy as the original model. 
    \item Using PaReprop scales better on throughput with memory than standard reversible backpropagation. In particular, PaReprop leads to more favorable memory vs. throughput trade-offs, i.e. our method can achieve higher throughput at any given threshold of memory used.
\end{enumerate}

\section{Related Works}
\paragraph{Reversible architectures} are a type of neural network architecture based on NICE~\cite{dinh2014nice, dinh2016density}, which was an early model for generative flow-based image generation~\cite{ho2019flow++, kingma2018glow}. Reversible ResNet~\cite{gomez2017reversible} is a type of reversible architecture that uses the NICE invertible transformations to enable memory-efficient image classification. Other researchers have built upon this idea by proposing improved reversible CNN models using ODE characterizations~\cite{chang2018reversible, li2021m, sander2021momentum}, momentum~\cite{li2021m, sander2021momentum}, and several other improvements~\cite{hascoet2019layer, finzi2019invertible, behrmann2019invertible, song2019mintnet}. Recently, reversible networks have also been adapted to core NLP tasks in Reformer~\cite{kitaev2020reformer} and to several core vision tasks in Rev-ViT~\cite{revvit}. Crucially, Rev-ViT~\cite{revvit}, under very strict parity constraints on parameters, FLOPs and activation sizes of the proposed models, shows reversible models to an equivalently powerful class of models as vanilla transformer but with the added benefit of extremely memory-efficient training. 

\paragraph{Extensions of reversible transformers} have been recently proposed that allow them to be used in more general settings. Rev-ViT and Rev-MViT were proposed in~\cite{revvit} as demonstrations that reversible architectures can be made from both isotropic and hierarchical transformers. Recently, work has also been done to show non-reversible checkpoints of models can be rewired to become reversible backbones for temporal action localization very efficiently, using much less compute than it would take to train a model from scratch~\cite{re2tal}. 
Similar work has also been done to extend these results to natural language transformers such as BERT, BART, and OPT~\cite{liao2023make}.
Our results adds to these works by additionally proposing Rev-Swin and Rev-RoBERTa models~\cite{swin, roberta}, as well as an orthogonal throughput improvement to all of these methods.

\section{\underline{Pa}rallelized \underline{Re}versible Back\underline{prop}agation}
We begin with a review of the reversible transformation (Section~\ref{subsec:rev_transform}), and subsequently introduce the vanilla memory-efficient reversible training algorithm (Section~\ref{subsec:revvit}) and its application to training modern transformers. Then, we present \textbf{PaReprop}, our proposed procedure for speeding up reversible backpropagation with parallelized activation and gradient computation (Section~\ref{subsec:parallelrevvit}). 

\subsection{Reversible Transformations}
\label{subsec:rev_transform}
A reversible transformation $T$ maps inputs $I_1$ and $I_2$ to outputs $O_1$ and $O_2$ in an invertible process even if there is no analytical inverse to the transformation. 
We will use intermediate functions $F$ and $G$, which need not be invertible. 

\begin{figure}[!tb]
    \centering
    \includegraphics[width=.45\textwidth]{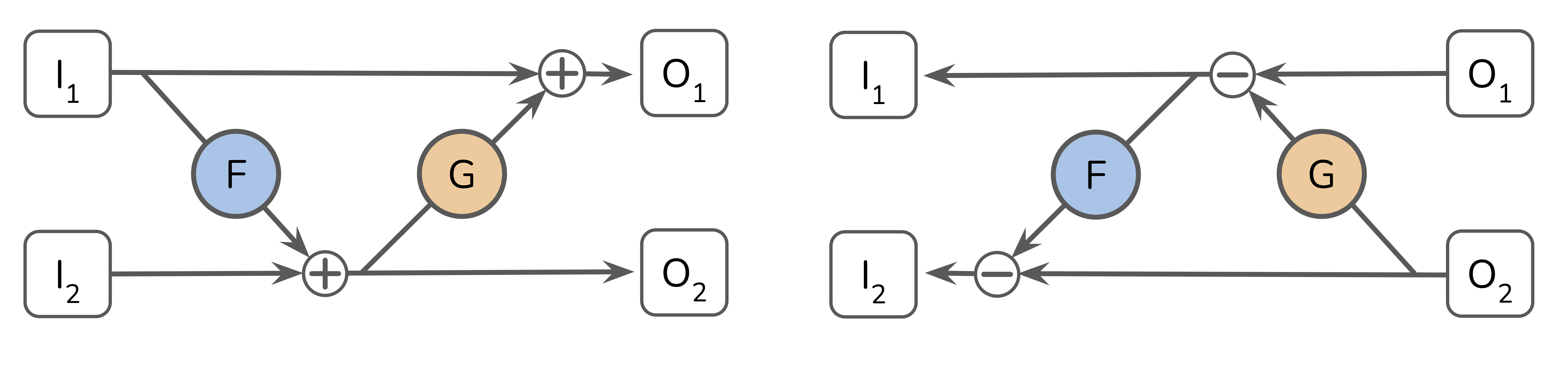}
    \caption{\textbf{Illustration of the Reversible Transformation} with arbitrary functions $F,G$ and the forward process (left) and backward process (right). See Eq.~\ref{eq:rev_transform} for the mathematical definition.}
    \label{fig:nice_flow}
\end{figure}

The reversible transformation applies each function $F$ and $G$ one at a time, first to $I_1$ and then to $I_2 + F(I_1)$, and adds it to the other input, providing a simple method to recover the inputs from the outputs.
This process is illustrated in Figure~\ref{fig:nice_flow}, and the final transformation is given by 

\begin{equation}
        \textbf{I} = \begin{bmatrix} I_1 \\ I_2 \end{bmatrix} 
        \xmapsto[T]{\quad}
        \begin{bmatrix} I_1 + G(I_2 + F(I_1)) \\ I_2 + F(I_1) \end{bmatrix}  
        = \begin{bmatrix} O_1 \\ O_2 \end{bmatrix}
        := \textbf{O} 
        \label{eq:rev_transform}
\end{equation}

It requires one call of $F(\cdot)$ and $G(\cdot)$ to compute either $T$ or an inverse $T'$, so both the forward and the backward pass require the same computational cost. 

\subsection{Reversible Transformers}
\label{subsec:revvit}

\begin{figure}[!t]
    \centering
    \includegraphics[width=0.47\textwidth]{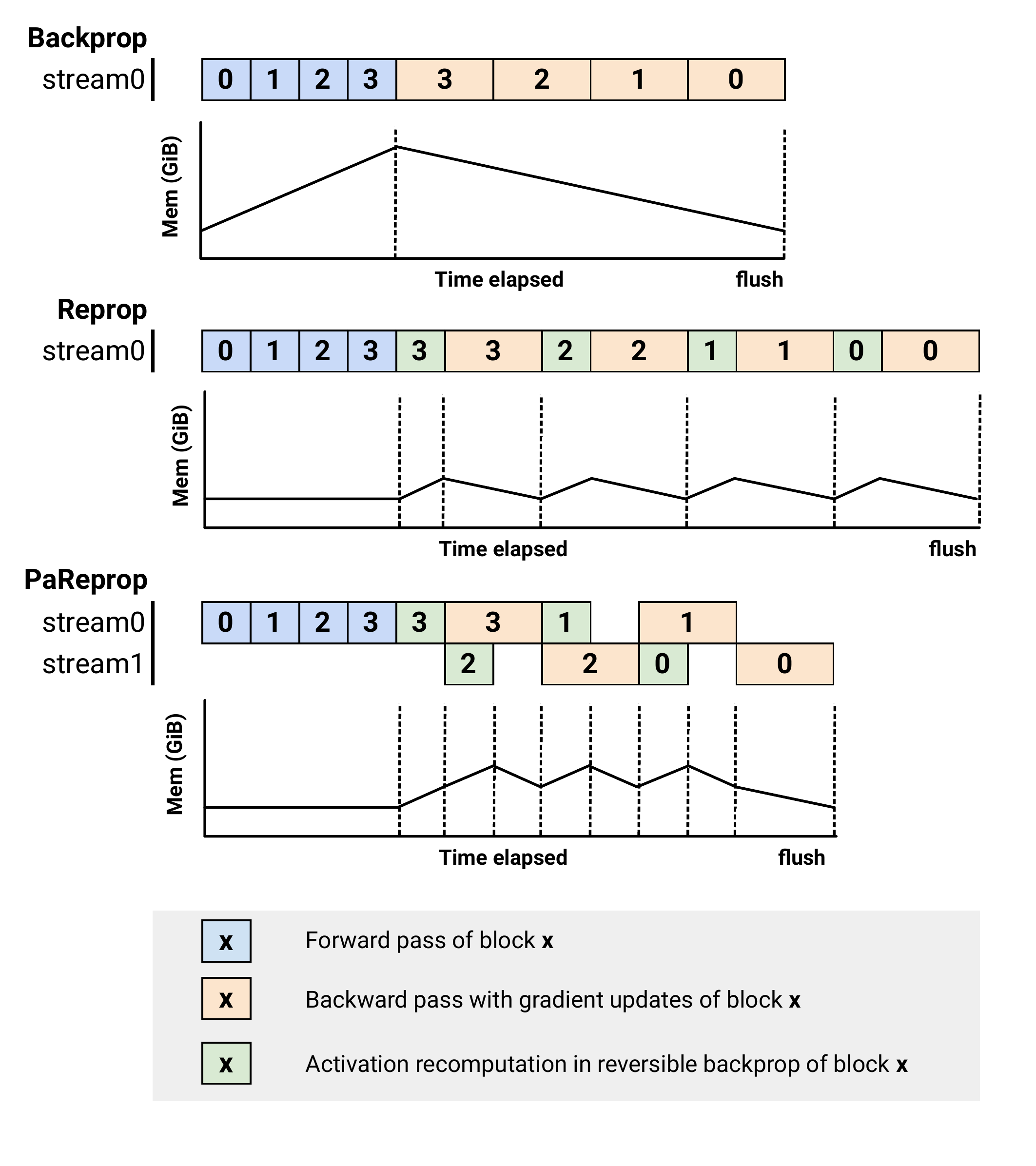}
    \caption{\textbf{Parallelized Reversible backpropagation}. PaReprop (bottom) parallelizes the activation re-computation stage of block $i-1$ (green blocks) with the gradient computation stage of block $i$. Reversible training drastically alleviates training memory burden of vanilla networks (top vs. middle rows) but introduces additional computational burden of activation re-computation in the backward pass (see \cite{revvit}). PaReprop further alleviates this re-computation burden, thereby making reversible architecture a practical choice for deep transformer training.}
    \label{fig:gpipe_diagram}
    \vspace{-10pt}
\end{figure}

Reversible Transformers ~\cite{gomez2017reversible, revvit} are a family of memory-efficient models based on these reversible transformations.
For their application, we will make use of the property that they can perfectly recompute any input $\mathbf{I}$ from its output $\mathbf{O}$, which can be used at the granularity of transformer blocks. 
For vision transformers, $F(\cdot)$ is set to be the attention block while $G(\cdot)$ is set to be the MLP block. 

\begin{figure}[!tb]
    \centering
    \includegraphics[width=0.47\textwidth]{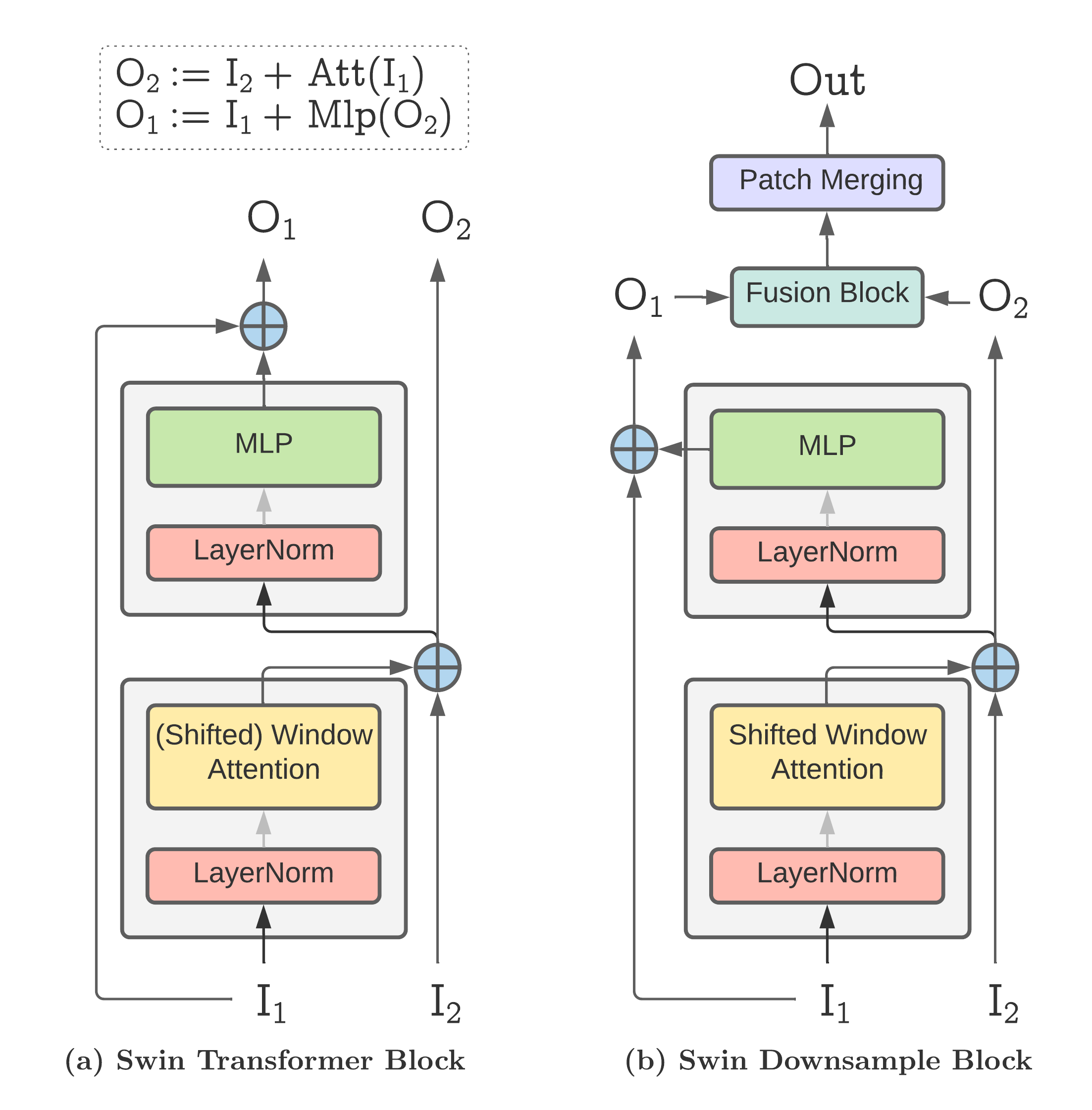}
    \caption{\textbf{Reversible Swin}. \cite{revvit} introduces the Reversible ViT and MViT architectures. Following the same principles, we introduce the reversible Swin architecture and benchmark all the three reversible architectures with PaReprop.  
    We showcase (a) a typical Swin block, as well as (b) a downsample block for processing at multiple hierarchies of scale.}
    \label{fig:revswin_diagram}
    \vspace{-00pt}
\end{figure}

\begin{figure*}[!tb]
    \centering
    \includegraphics[width=.95\textwidth,height=0.52\textheight]{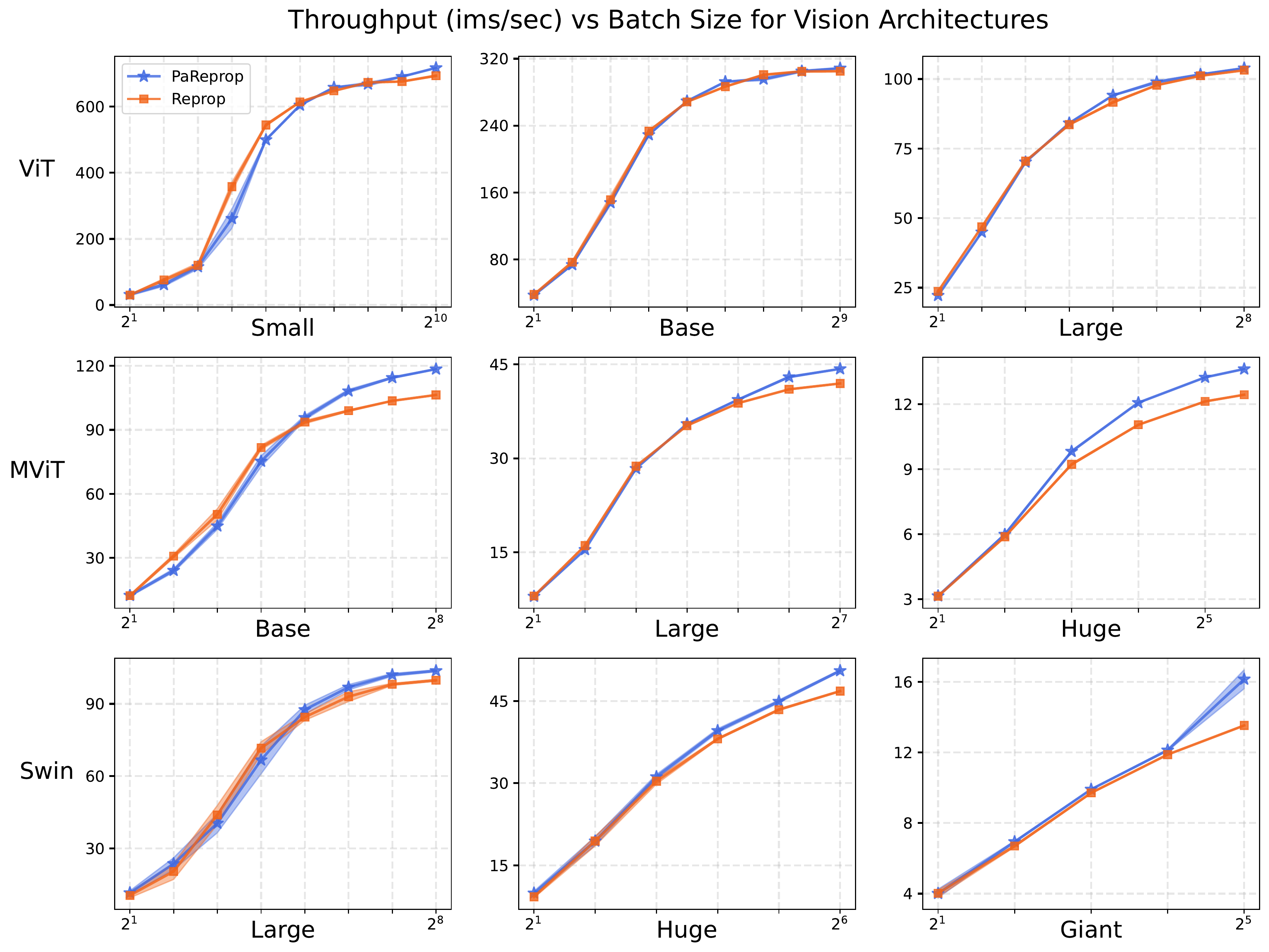}
    \caption{\textbf{PaReprop Vision Training throughput vs. Batch size} across model architectures and sizes. $1 \sigma$ error bars are shown. Accuracies are kept intact with PaReprop (Section~\ref{sec:results}). }
    \label{fig:all_bb_thr}
    \vspace{-10pt}
\end{figure*}

This means that during our forward pass, we do not need to store any activations. 
In the backward pass, we recompute the activations of the current block using the output, and then backpropagate to recover our gradients and update our weights. 
We then delete the activations of the current block, and repeat this process for the next block in normal Reversible Backpropagation, or Reprop, in Figure~\ref{fig:gpipe_diagram}.

However, the backward pass of block $N$, i.e. the gradient update, is not needed to recompute the activations of block $N-1$. 
Therefore, if we can hide the forward pass of block $N-1$ within the backward pass of block $N$, we can theoretically speedup our computation on par with that of normal backprop.
This is our key insight, which we will now present in detail.

\subsection{Parallelizing with Two Streams}
\label{subsec:parallelrevvit}

Our method's key contribution is both theoretical and practical. 
The first is illustrated in Figure~\ref{fig:gpipe_diagram}, where our PaReprop method is able to parallelize the backward pass of a normal reversible backprop so that it takes nearly the same time as vanilla backprop. 
We do this by performing the gradient update for block $N$ at the same time as the activation recomputation for block $N-1$, as there is no dependence once we have the activations for block $N$. 
Approximating a backward pass as twice the time as a forward pass, this means theoretically we can hide 25\% of the computations and see this speedup in our throughput.

However, achieving this parallelization is rather tricky. Standard autodifferentation frameworks like PyTorch use a CUDA stream by default to maintain sequential ordering for GPU operations. 
Our method extends this by maintaining multiple streams to parallelize operations over.
However, these streams enforce that forward and backward passes occur on the same stream, which causes issues if we implement PaReprop naively by keeping one stream for the activation recomputation and another for the gradient updates.
This necessitates our alternative computation scheme.

Another problem is that PyTorch is unable to free memory efficiently in parallel processes as there is no asynchronous implementation of CUDA free yet. Thus, it requires a costly CUDA free operation which synchronizes our streams and thus slows our process down significantly. In practice, it's most beneficial to run our PaReprop method at anywhere from 33\% to 50\% of the empirical maximum batch size so that we don't hit this trap.

\subsection{Reversible Swin and RoBERTa}
\label{subsec:revmodels}

In order to demonstrate the generality of our method to other architectures, we propose two novel reversible architectures based on established models: Swin Transformer and RoBERTa.
Figure \ref{fig:revswin_diagram} shows our modifications to the original Swin architecture. 
We follow suit with the original reversible ViT authors and choose to keep the Attention blocks and MLP blocks as our reversible subfunctions. 
We also demonstrate how we handle multiscale features in the architecture by utilizing a fusion block (either averaging or a simple MLP) before the usual patch merging layer. 

\section{Results}
\label{sec:results}

In this section, we present our experimental results of our proposed method, denoted as PaReprop, in comparison with the original reversible ViT, denoted as Reprop. 
We analyze our method over the choice of backbone size (from 5.5M to over 2B parameters), architecture class (ViT~\cite{vit}, Swin~\cite{swin}, MViT~\cite{mvit, mvitv2}, RoBERTa~\cite{roberta}), data modalities (Vision and NLP), GPU training memory allocated, input sequence length (for RoBERTa) and batch size. The primary metric we are concerned with is training throughput (images/sec or sequence/sec), which is the number of images (or sequences) we can process per second, as our measure of speed. We benchmark on 224$\times$224 image classification and standard text classification.

As shown in \cite{revvit}, the reversible models attains the same accuracy as the original reversible ViT, and we do not change the underlying algorithm but simply propose a faster implementation, so we focus instead on analyzing the speedup of our method instead of performance.
All of our results are run on a single NVIDIA A100 GPU with 80GB of memory w/ AdamW as our optimizer. Our results were similar on an NVIDIA RTX 2080 Ti as well as with SGD.

\begin{figure}[!t]
    \centering
    \includegraphics[width=0.45\textwidth]{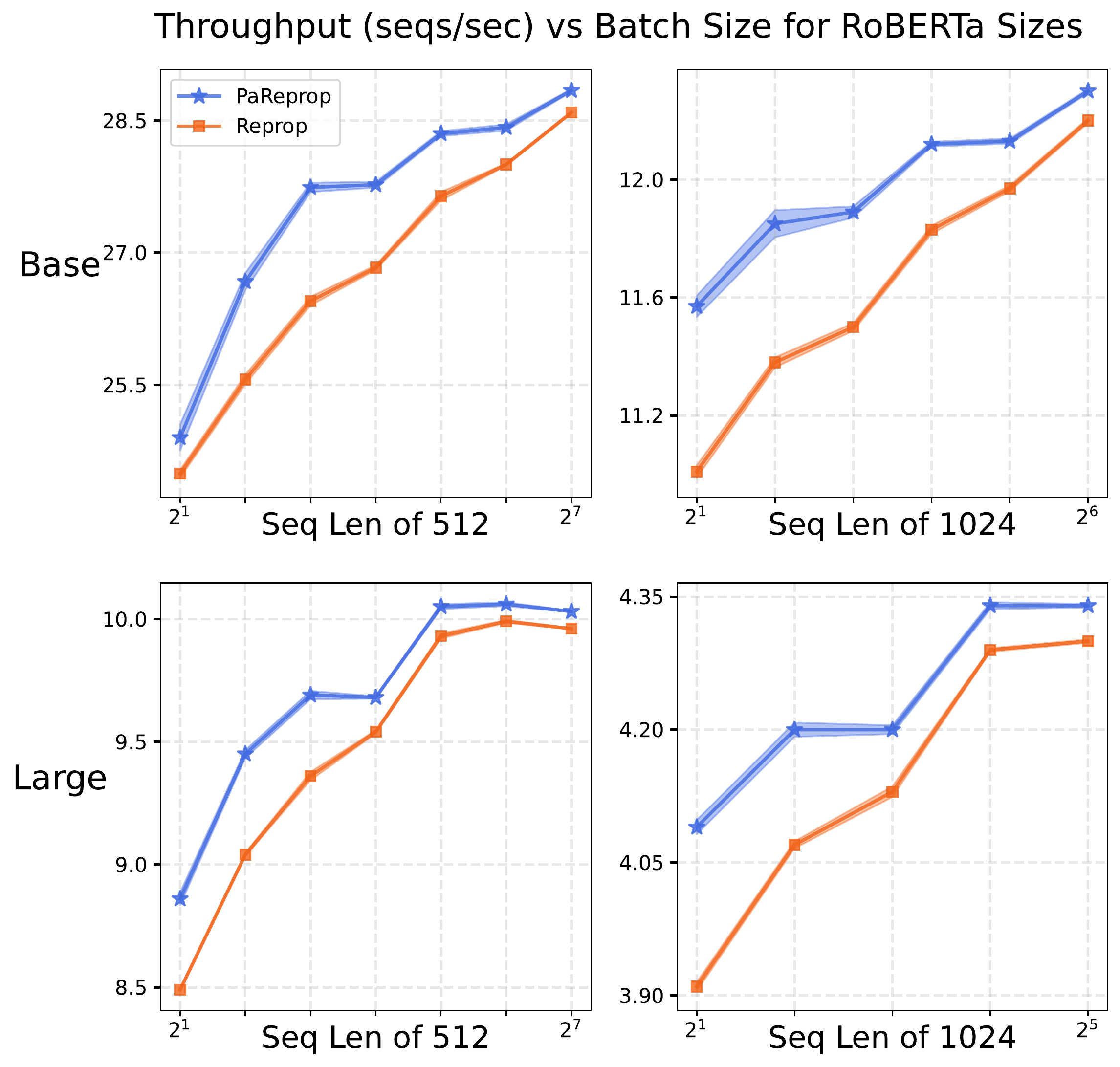}
    \caption{\textbf{PaReprop NLP Training Throughput across Sequence length and Batch Sizes}. Comparison of our method on RoBERTa, a language transformer. One s.d. error bars are shown.}
    \label{fig:roberta_thr}
    \vspace{-15pt}
\end{figure}

\subsection{PaReprop Training Throughput is Better}
\label{subsec:s2_better}
In the first experiment, we compare our PaReprop method with the original Reprop method used in the original reversible ViT. 
We compare the top throughput achieved over batch sizes of $\{1, 2, 4, \dots, 256, 1024\}$ (as allowed by memory) for each backbone. 
In these cases, we run on standard image classification, but our results will hold over any choice of task (video understanding, etc.). 
We see that across three different architecture families (ViT, MViT, Swin) and three choices of model sizes, our method outperforms Reprop, in some cases drastically. 

\vspace{-8pt}
\paragraph{Vision Transformers} show a mostly matched throughput between PaReProp and standard reversible backpropagation (Figure~\ref{fig:all_bb_thr}, top row). We find that because ReProp
can already utilize a large batch size, the GPU utilization is quite high and the PaReprop kernels are unable to run in parallel. 

\vspace{-8pt}
\paragraph{Hierarchical Transformers} enjoy a much more pronounced benefit with PaReprop such as Multiscale Vision Transformer (Figure~\ref{fig:all_bb_thr}, middle row) and Swin Transformer (Figure~\ref{fig:all_bb_thr}, bottom row; Figure~\ref{fig:revswin_topthr}).
Hierarchical transformers have non homogeneous architectures that allow PaReprop to hide the recomputation latency. They have several small kernels that can be parallelized more effectively, which leads to significant speedups of 9.8\% on the largest MViT-H and a 19.3\% increase on largest Swin-G, consistently outperforming the standard reversible backpropagation method (Reprop). 
This shows that PaReprop provides significant speedup for vision-specific models.

\vspace{-8pt}
\paragraph{NLP Transformers.} Finally, we also clearly demonstrate PaReprop gains on the natural language modality. Reversible models have also successfully been applied to language tasks such as Reformer.  Following \cite{revvit}, we extend RoBERTa \cite{roberta} to Rev-RoBERTa and provide throughput benchmarking results on sentiment analysis. Note that as shown in Rev-ViT~\cite{revvit} using a simple reversible rewiring of the model  maintains the original accuracy of the vanilla model. 
Figure~\ref{fig:roberta_thr} shows our method outperforming the original Reprop by large amounts across both choices of model size (Base and Large) and sequence lengths (512 and 1024). 

\begin{figure}[!tb]
    \centering
    \includegraphics[width=0.47\textwidth]{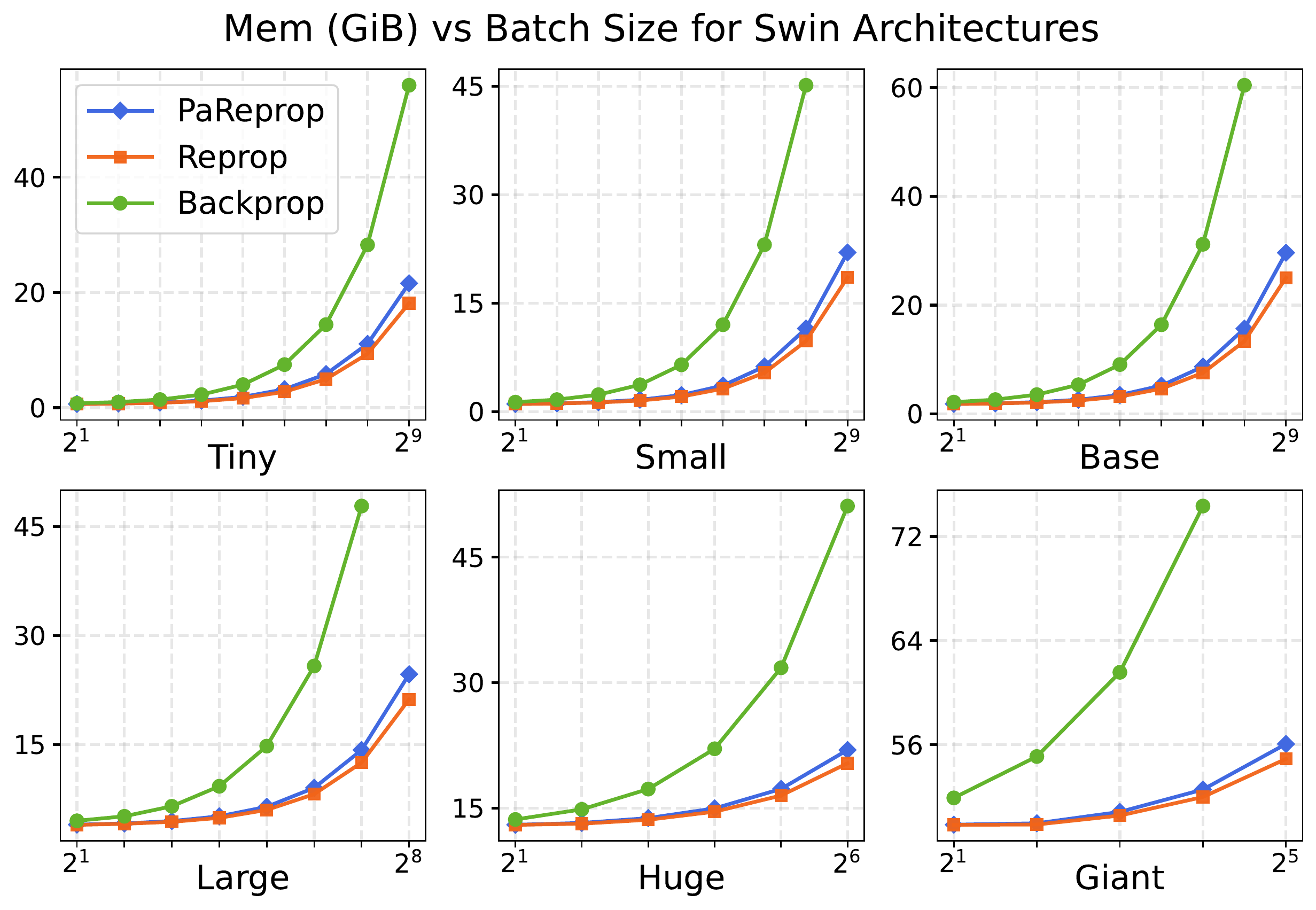}
    \caption{\textbf{PaReprop Memory Vs. Batchsize}. Our method, PaReprop , as well as standard reversible backprop, Reprop, use comparatively much less memory than nonreversible approaches (Backprop). The memory increase to achieve proposed speedups is negligible compared to the overall savings.}
    \vspace{-20pt}
    \label{fig:revswin_mem}
\end{figure}

\subsection{Training GPU Memory Trends}

We also investigate the effect of memory on our method (Figure~\ref{fig:revswin_mem}). Specifically, we compare the memory used by our method with the original Reprop and the vanilla backprop, and show that our method is more memory efficient.
As shown in the plots, using any kind of reversible backpropagation offers memory savings of up to almost 3x in some cases, which allows us to significantly extend the batch sizes that we can use. 
In these scenarios, using our parallelized reversible backpropagation requires only an extra fraction of the small amount of memory being used to maintain parallelization and allows us to achieve higher throughputs. This finding is consistent across all model architectures we tested, but we illustrate most of our findings on Swin in Figure~\ref{fig:revswin_mem} for simplicity.

\section{Conclusion}

We present Parallelized Reversible Backpropagation (PaReprop), a fast reversible backpropagation algorithm for training reversible transformer in both Vision (Rev-ViT, Rev-Swin, Rev-MViT) and NLP (Rev-RoBERTa). PaReprop parallelizes the backward activation recomputation, an additonal overhead introduced in the reversible networks, with the gradient computation itself in backpropgation.
By allowing recomputation to overlap with gradient calculation, the additional latency of recomputation is hidden thereby increasing the training throughput while still maintaining extremeley memory-efficient training. 
Our method maintains the same accuracy as the vanilla method, and achieves better throughput on all of the models we tested, outperforming some by up to 20\%. 

{\small
\bibliographystyle{ieee_fullname} 
\bibliography{refs} 
}

\end{document}